\documentclass[sigconf]{acmart}
\pdfoutput=1
\usepackage{booktabs} 
\usepackage[normalem]{ulem}
\usepackage{float}
\usepackage{enumitem}
\copyrightyear{2019}
\usepackage{amsmath}
\usepackage{algorithm}
\usepackage[noend]{algpseudocode}
\usepackage{graphicx}
\usepackage{graphics}
\usepackage{caption}
\usepackage{subcaption}

\begin{document}
\title{IVO: Inverse Velocity Obstacles for Real Time Navigation }

\author{P. S. Naga Jyotish}
\authornote{Both authors contributed equally to this research.}
\email{srisai.poonganam@research.iiit.ac.in}
\affiliation{%
  \institution{Robotics Research Center, IIIT Hyderabad}
}

\author{Yash Goel}
\authornotemark[1]
\email{ygoel@me.iitr.ac.in}
\affiliation{%
  \institution{Robotics Research Center, IIIT Hyderabad}
}

\author{A. V. S. Sai Bhargav Kumar}
\email{vseetharam.a@research.iiit.ac.in}
\affiliation{%
  \institution{Robotics Research Center, IIIT Hyderabad}
}

\author{K. Madhava Krishna}
\email{mkrishna@iiit.ac.in}
\affiliation{%
  \institution{Robotics Research Center, IIIT Hyderabad}
}

\begin{abstract}
In this paper, we present \textbf{IVO: Inverse Velocity Obstacles} an ego-centric framework that improves the real time implementation. The proposed method stems from the concept of velocity obstacle and can be applied for both single agent and multi-agent system. It focuses on computing collision free maneuvers without any knowledge or assumption on the pose and the velocity of the robot. This is primarily achieved by reformulating the velocity obstacle to adapt to an ego-centric framework. This is a significant step towards improving real time implementations of collision avoidance in dynamic environments as there is no dependency on state estimation techniques to infer the robot pose and velocity. We evaluate IVO for both single agent and multi-agent in different scenarios  and show it's efficacy over the existing formulations. We also show the real time scalability of the proposed methodology.  
\end{abstract}



\maketitle

\section{Introduction}

Autonomous navigation has gained a lot of attention in the recent years. They find applications in the fields like self-driving cars, crowd simulations, rescue operations, payload transferring etc. All these applications require a collision avoidance scheme for a safe navigation of the system to the goal. There have been quite  a few approaches like \cite{van2008reciprocal}\cite{van2011reciprocal} which present collision avoidance schemes but are computationally complex due to the non-convex nature of collision avoidance constraint. Also these schemes generally estimate whether the agent is on collision course with the other participants based on the states of the agent and the participants. A slight variance in the state estimation can lead to false detection which keeps propagating and can lead to system failure. In this paper, we present a novel methodology for collision avoidance that removes the reliance on the state of the robot. Our approach stems from the concepts of Velocity Obstacle \cite{fiorini1998motion} and ego-centric motion planning. 

\subsection{Contribution and Main Results}

The principal contribution of the present work is the construction of efficient collision avoidance scheme for autonomous navigation called Inverse Velocity Obstacles (IVO). Our approach is a variant of Velocity Obstacle method presented in \cite{fiorini1998motion}, which is widely used technique for collision avoidance in a dynamic environment. Our method inherits all the salient features and incorporates capability to handle the uncertainty in collision detection that occur due to the error in state estimation. This is achieved by implementing the algorithm in an ego-centric framework. Due to the very nature of the implementation, it can be easily extended to multi-agent collision avoidance problem by implicitly assigning each agent with the same collision avoidance scheme. We also show that the low computational complexity and lower noise in collision detection of the approach significantly improves the chances for real time implementations as there is dependency on the state estimation techniques for inferring the self states of each agent.

On implementation side, we show the efficacy of Inverse Velocity Obstacles method by evaluating it in various scenarios for both single and multi-agent systems. Our simulations show that even for the agents as high as 50 can generate safe motions. We also show the variance of false collision detection is reduced significantly compared to a Velocity  Obstacle approach. We have also show the real time  potential of the  presented approach  by implementing it on real drone and also the approach can be easily parallelized as each agent computation is independent.

\subsection{Layout of the paper}
The rest of the paper is organized as follows, Section \ref{rel_work} presents a brief overview of the previous works. Section \ref{back} reviews the concepts of Velocity Obstacle. In Section \ref{IVO} we present our approach, Inverse Velocity Obstacles and derive its formulation. Section \ref{Nav_Agents} describes the implementation details for the navigation of single and multi-agent systems. In Section \ref{res} we evaluate our method in different scenarios and demonstrate the performance in real time. We conclude our work in Section \ref{concl}.

\section{Related Work}\label{rel_work}
In this section, we present an overview of the approaches on collision avoidance and navigation in dynamic environment. Quite a few approaches \cite{borenstein1991vector}, \cite{faverjon1987local}, \cite{fox1997dynamic}, \cite{kanehiro2008local} assume that the obstacles are static and plan for the control to avoid the collision. In case of moving obstacles these replan based on the updated positions of the obstacles. But these fail to generate the safe trajectories around the fast moving obstacles. In \cite{fulgenzi2007dynamic}, \cite{de1994avoidance}, \cite{hsu2002randomized}, \cite{martinez2009collision} the future position of obstacles are computed by extrapolating with the current velocity to handle high velocities. But these approaches cannot handle the reactive nature of the other agents. Many works like \cite{pettre2006real}, \cite{treuille2006continuum}, \cite{sud2008real}, \cite{gayle2007reactive} have focused on crowd simulation in which each agent considers the other agents as obstacles and navigates independently.   

 Centralized planning scheme on a given configuration space in the case of multiple agents is presented in \cite{lavalle1998optimal}, \cite{sanchez2002using}. These works majorly focus on optimal coordination and cannot be scaled up for real time implementation.  A method called Velocity Obstacle based on velocity is presented in \cite{fiorini1998motion} for moving obstacles which provides the robot a condition to avoid collision with obstacle with a known velocity. A variant called  Recursive Velocity Obstacles \cite{kluge2004reflective} is proposed, which considers the reactive behaviour of the other participants. However, this approach leads to the oscillations of the agents which sometimes may not converge. To address issue a extension to the Velocity obstacle called Reciprocal Velocity Obstacle (RVO)\cite{van2008reciprocal} is presented, where both the agents which are on the course of collision select the velocities that bring them outside the RVO which is generated by the other agent. But this requires the knowledge of current pose and velocity of the obstacle which might bottleneck the update rates during real time implementation. They are several other extensions of Velocity Obstacle like \cite{singh2013reactive}\cite{kumar2018novel}.
 
 To address this in this paper, we present an ego-centric based framework called Inverse Velocity Obstacles (IVO), which does not require the knowledge of robot's pose and velocity. This eliminates the state estimation layer reducing the computational time (for state estimation) and false collision detection which aids in real time implementation.

\section{Preliminaries}\label{back}

\subsection{Velocity Obstacle}
In this section, we briefly review the original concept of Velocity Obstacle  and analyze its behaviour in in the presence of state, actuation and perception uncertainties.

\subsubsection{Definition}
\begin{figure}[h]
    \includegraphics[page=1,width=1\linewidth]{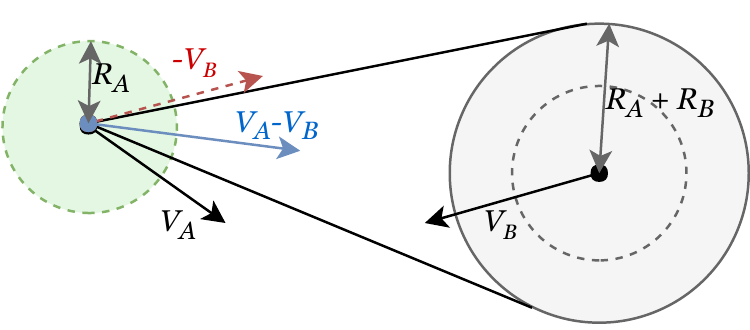}
    \caption{Velocity obstacle for agent $A$ induced by obstacle $B$}
    \label{fig:VO}
\end{figure}

Consider a mobile robot (our agent) and an obstacle, both taking the shape of a disc of radius $R_A$ and $R_B$ respectively, be denoted by $A$ and $B$. The velocity obstacle for robot $A$ induced by obstacle $B$, denoted by $VO_{A|B}$, is the set of velocities of $A$ which can result in a collision with $B$ at some point in the future. Let $C_A$ and $C_B$ represent the centres of $A$ and $B$ respectively. The robot and obstacle are geometrically modified such that the robot takes the form of a point object and the obstacle grows its radius to $R_A+R_B$. If $B$ is a static obstacle, a cone can be constructed with the vertex on $A$ and the edges touching $B$ as shown in the figure \ref{fig:VO}. This cone represents the set of velocities of $A$ which lead to a collision. In case the obstacle is in motion, it is assumed to be static by taking a relative velocity of $A$.

\subsubsection{Implementation problems}
The obvious assumption from the definition of the velocity obstacle is that we need to track the velocity of the robot along with the position and velocity of the obstacle. In case of planning trajectories on a global frame, we also need to track the positions of robot and obstacle with respect to a global frame. Though we can plan trajectories in robot's frame, this still needs us to have an estimation of the velocity of the robot. Generally, we take the instantaneous velocity from a sensor. This accounts for an additional noise in estimation of the velocity of the robot apart from the noise we end up having in the states of the obstacle. Other prominent methods include state estimation using SLAM which is not as reliable as the feed from the sensor since SLAM methods tend to break when complex maneuvers are involved.

\section{Inverse Velocity Obstacle} \label{IVO}
In this section, we propose a new concept of "Inverse Velocity Obstacle" to minimize the uncertainty in collision detection during the planning phase. This integrates into our optimization framework which provides controls leading to collision free and smooth trajectories.

\subsection{Definition}
\begin{figure}
    \centering
    \includegraphics[page=1,width=1\linewidth]{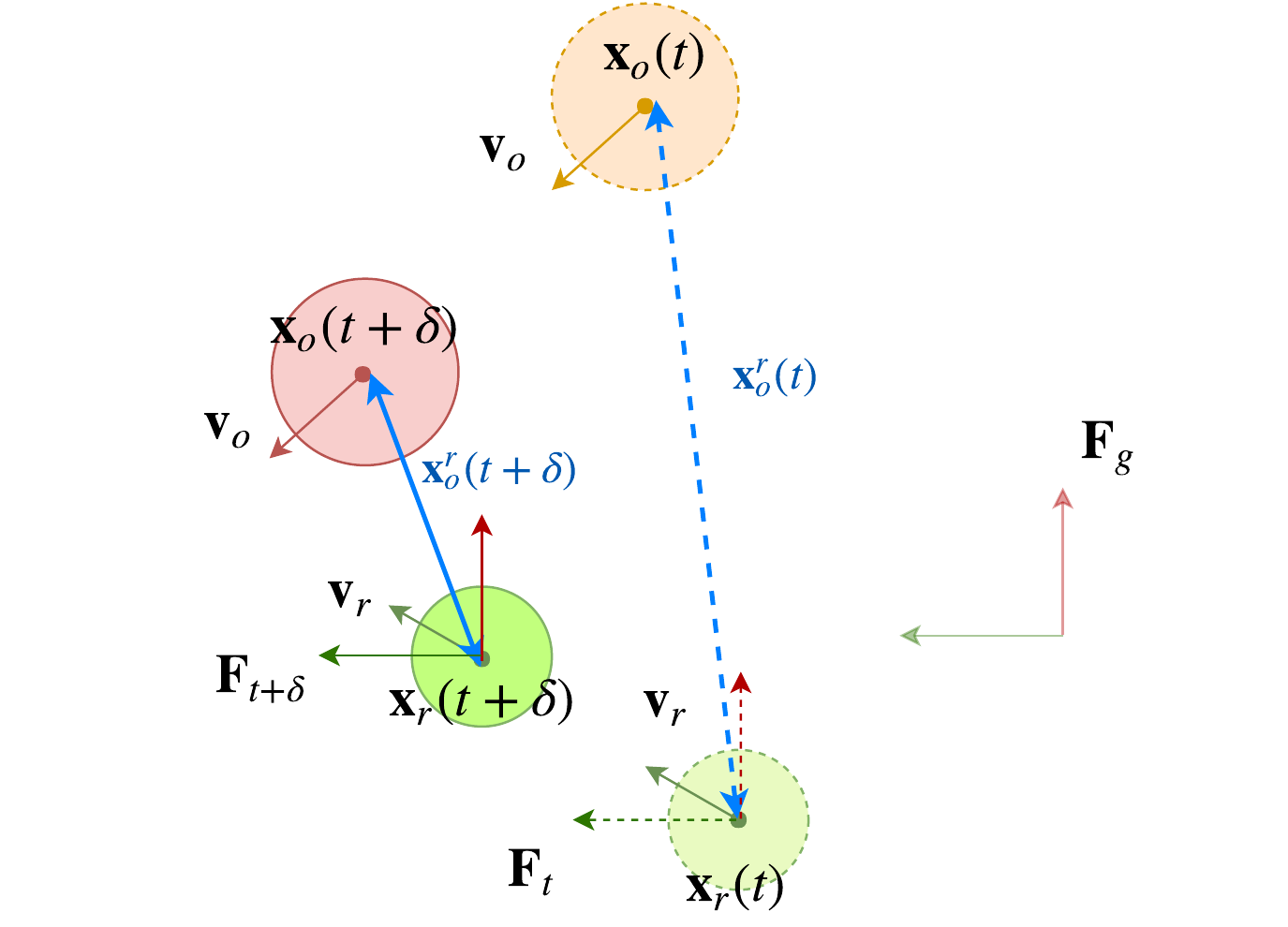}
    \caption{$\textbf{x}_r$, $\textbf{v}_r$ denote the position and velocity of the agent while $\textbf{x}_o$ and $\textbf{v}_o$ denote the position and velocity of the obstacle in global frame. $\textbf{x}^r_o$ and $\textbf{v}^r_o$ denote the position and velocity of the obstacle as seen from the agent's frame (agent is at origin and stationary in this frame).}
    \label{fig:ivo-explanation}
\end{figure}

The idea is simple - Instead of assuming that the obstacle is stationary, we assume that the robot is stationary and get a relative velocity vector for the obstacle. At this point, our robot is stationary at the origin (since we are in an ego-frame). We also make the obstacles point objects and grow the radius of the robot to $R_A+R_B$. Now, we find a relative velocity for our robot (which is stationary in the relative frame) which is outside the collision cone. A simple case is demonstrated in the figure \ref{fig:ivo-explanation}, where $\textbf{x}_i(t) = [x_i(t) \ \ y_i(t)]^T$ and $\textbf{v}_i(t) = [\dot{x_i} \ \ \dot{y_i}]^T$. We show that we can calculate the relative velocity of the obstacle as seen by the agent using the ego-centric observation of the obstacle by the agent at two consecutive time instance, here \(t\) and \(t + \delta\), as shown in \ref{eq:rel-velo}

\begin{equation*}
    \begin{bmatrix}
    \dot{x^r_o} \\ \dot{y^r_o}
    \end{bmatrix} = \begin{pmatrix}\begin{bmatrix}
    x^r_o(t+\delta) \\ y^r_o(t+\delta)
    \end{bmatrix} - \begin{bmatrix}
    x^r_o(t) \\ y^r_o(t)
    \end{bmatrix}\end{pmatrix}/\delta
    \label{eq:rel-velo}
\end{equation*}

For any time instance, \(t\) suppose the global position of the obstacle moving with velocity \(\textbf{v}_o\) and agent moving with velocity \(\textbf{v}_r\) be \(\textbf{x}_o(t)\) and \(\textbf{x}_r(t)\) respectively. At the next time instance, the global positions of the obstacle and agent will be \(\textbf{x}_o(t+\delta)\) and \(\textbf{x}_r(t+\delta)\) respectively. The ego-centric observations of the obstacle by the agent for these instances is \(\textbf{x}^r_o(t)\) and \(\textbf{x}^r_o(t+\delta)\) for agent frame \(\textbf{F}_t\) and \(\textbf{F}_{t+\delta}\) respectively.

So, the global position of the obstacle at first instance is

\begin{equation*}
\textbf{x}_o(t) = {^g_t}{\textbf{T}}.{\textbf{x}}{^r_o}(t)
\end{equation*}
\begin{equation*}
\textbf{x}_o(t) = {\textbf{x}}{^r_o}(t) + {\textbf{x}}{_r}(t)
\end{equation*}

Similarly for the second instance we have
\begin{equation*}
\textbf{x}_o(t+\delta) = {^g_{t+\delta}}{\textbf{T}}.{\textbf{x}}{^r_o}(t+\delta)
\end{equation*}
\begin{equation*}
\textbf{x}_o(t+\delta) = {\textbf{x}}{^r_o}(t+\delta) + {\textbf{x}}{_r}(t+\delta) \\
\end{equation*}
\begin{equation*}
\textbf{x}_o(t+\delta) = {\textbf{x}}{^r_o}(t+\delta) + {\textbf{x}}{_r}(t) + \textbf{v}_r*\delta \\
\end{equation*}

Therefore the obstacle velocity in the global frame is
\begin{equation*}
\textbf{v}_o = (\textbf{x}_o(t+\delta) - \textbf{x}_o(t))/\delta 
\end{equation*}
\begin{equation*}
\textbf{v}_o = ({\textbf{x}}{^r_o}(t+\delta) - {\textbf{x}}{^r_o}(t) + \textbf{v}_r*\delta)/\delta \\
\end{equation*}

And hence the relative velocity of the obstacle with respect to the agent is
\begin{equation*}
    \textbf{v}{_o^r} = \textbf{v}_o - \textbf{v}_r
\end{equation*}
\begin{equation*}
    \textbf{v}{_o^r} = ({\textbf{x}}{^r_o}(t+\delta) - {\textbf{x}}{^r_o}(t) + \textbf{v}_r*\delta)/\delta - \textbf{v}_r
\end{equation*}
\begin{equation*}
    \textbf{v}{_o^r} = ({\textbf{x}}{^r_o}(t+\delta) - {\textbf{x}}{^r_o}(t))/\delta
\end{equation*}

Now, we write the collision cone using inverse velocity obstacles,

\begin{equation}
    f = \frac{(\textbf{r}^T \textbf{v})^2}{||\textbf{v}||^2} - ||\textbf{r}||^2 + (R_A+R_B)^2
    \label{eq:vo}
\end{equation}

\begin{equation*}
    \textbf{r} = \begin{bmatrix}
        x^r_o(t) \\
        y^r_o(t)
    \end{bmatrix},
    \textbf{v} = \begin{bmatrix}
        \dot{x^r_o(t)} \\
        \dot{y^r_o(t)}
    \end{bmatrix}
\end{equation*}









\section{Navigating agents} \label{Nav_Agents}

\subsection{Single Agent}
Let us start with the case of a single agent that follows a holonomic motion model and obstacles that do not have a complex behaviour but move with some constant velocity. Now, consider the following optimization with variables as $\textbf{u} = [u_x \ \  u_y]^T$ which represent the controls to the agent at a time instant $t$. The goal position in the agent's frame is denoted by $\textbf{g}^r$ and $\textbf{u}$ is the control given to the agent, which in this case is the change in the velocity. $\textbf{r}$ and $\textbf{v}$ represent the position and velocity of the obstacle as seen by the agent. The smoothing factor $\lambda$ can be adjusted based on the requirement. Let us assume that the maximum attainable velocity of the agent is $\textbf{v}_{max}$.

\begin{subequations}
    \label{eq:single-agent-goal}
    \begin{equation}
        \min_{u_x, u_y} J = ||\textbf{v}_{desired} - (\textbf{v}_{r}+\textbf{u})||^2 + \lambda||\textbf{u}||^2
    \end{equation}
    \begin{equation*}
        \textbf{v}_{desired} = \frac{\textbf{g}^r}{|\textbf{g}^r|}*v_{max}
    \end{equation*}
    \begin{equation}
        f(.) \leq 0: \frac{(\textbf{r}^T \textbf{v})^2}{||\textbf{v}||^2} - ||\textbf{r}||^2 + (R_A+R_B)^2 \leq 0
    \end{equation}
    \begin{equation*}
        \textbf{g}^r = \begin{bmatrix}
            g^r_x\\
            g^r_y
        \end{bmatrix}, u = \begin{bmatrix}
            u_x\\
            u_y
        \end{bmatrix}
    \end{equation*}
    \begin{equation*}
        \textbf{r} = \begin{bmatrix}
            x^r_o(t) \\
            y^r_o(t)
        \end{bmatrix},
        \textbf{v} = \begin{bmatrix}
            \dot{x^r_o(t)} - u_x \\
            \dot{y^r_o(t)} - u_y
        \end{bmatrix}
    \end{equation*}
\end{subequations}

The collision avoidance constraint, $f(.)$, exists for every possible pair of agent and obstacle within the sensor range of the agent. In section \ref{section:results-single-agent}, we experimentally show that this formulation is valid and the agent successfully avoids the obstacles and reaches the goal.

\subsection{Multiple Agents}
Let us consider $n$ agents that use the optimization routine mentioned in equation \ref{eq:single-agent-goal}. In this case, the obstacles may not necessarily move with constant velocity. For the sake of simplicity, we assume that every agent moves with some instantaneous velocity $dt$. Now, we scale the single agent problem to $n$ agents by considering every other agent to be an obstacle. Following this idea, a navigation algorithm for multi-agent scenario is described in Algorithm \ref{algo:multiagent}.

\begin{algorithm}
    \caption{Controls for agents in multi-agent setup}\label{algo:multiagent}
    \begin{algorithmic}
        \For{$i = 1$ to $n$}
            \For{obstacle, $j$, in obstacles in sensor range}
                \State $x^r_j(t) \gets \text{Position of an obstacle in agent's frame}$
                \State $\dot{x^r_j}(t)\gets (x^r_j(t) - x^r_j(t-dt)) \cdot dt$
                \State $R_j \gets \text{Radius of the obstacle}$
                \State $c_{avoid}(j) \gets f(x^r_j(t), \dot{x^r_j}(t), R_j)$
            \EndFor
            \State $\textbf{u}_i \gets \min\limits_{u_x, u_y} J$
        \EndFor
        \Return $\textbf{u} = [\textbf{u}_1 \ \textbf{u}_2 \ldots \textbf{u}_n]^T$
    \end{algorithmic}
\end{algorithm}

In section \ref{section:results-multiagent}, we experimentally show that the algorithm works for multiple agents with large values of $n$.

\section{Experimental Results}\label{res}
To evaluate the performance of the presented methodology we have tested in both single agent and multi agent scenarios. All the simulations are performed on Intel i7 processor @
3.2 GHz clock speed. The methodology is also validated on a real quadrotor. For this we used Parrot Bebop2. The detailed videos of all the simulations and real time implementations are available at [\href{https://sites.google.com/view/inverse-velocity-obstacle/}{this link}].

\subsection{Single agent}\label{section:results-single-agent}
\begin{figure}
     \centering
     \hspace{-0.2cm}
     \begin{subfigure}[b]{0.5\linewidth}
         \centering
         \includegraphics[width=\textwidth]{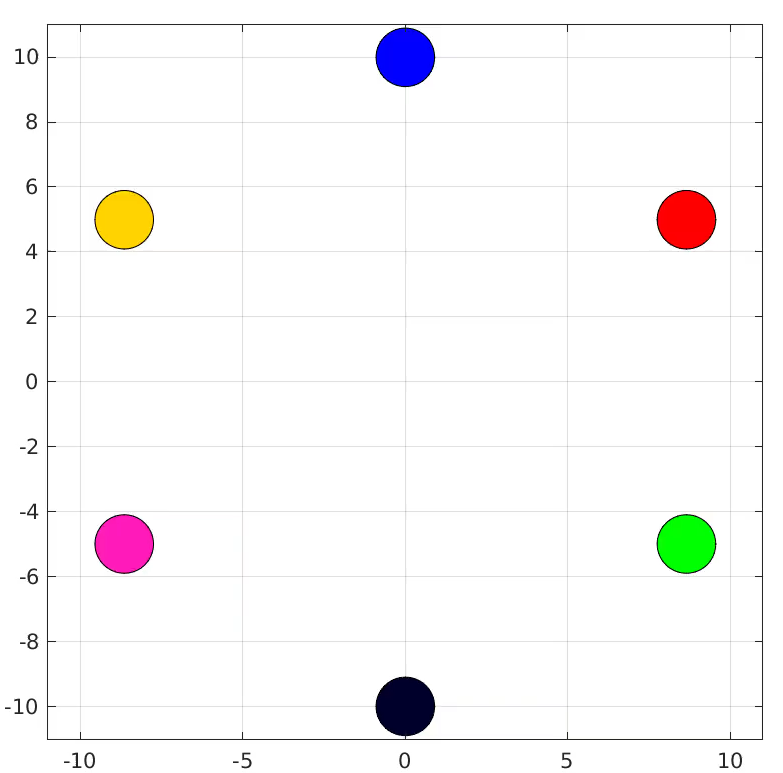}
         \label{SA:a}
     \end{subfigure}
     \begin{subfigure}[b]{0.505\linewidth}
         \centering
         \includegraphics[width=\textwidth]{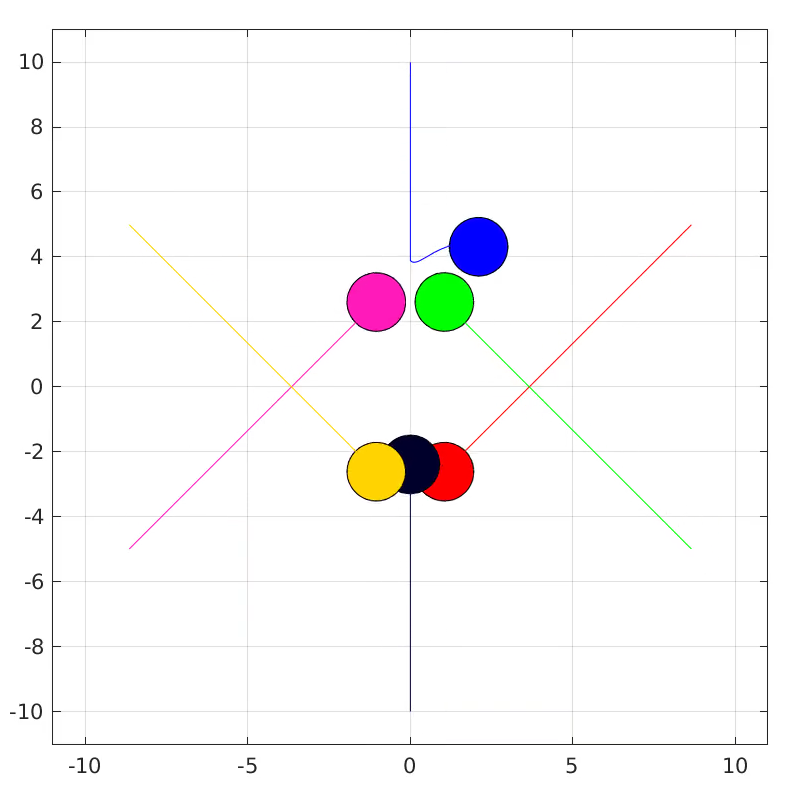}
     \end{subfigure}\label{SA:b}
     \vspace{0.1cm}
     \hspace{-0.2cm}
     \begin{subfigure}[b]{0.5\linewidth}
         \centering
         \includegraphics[width=\textwidth]{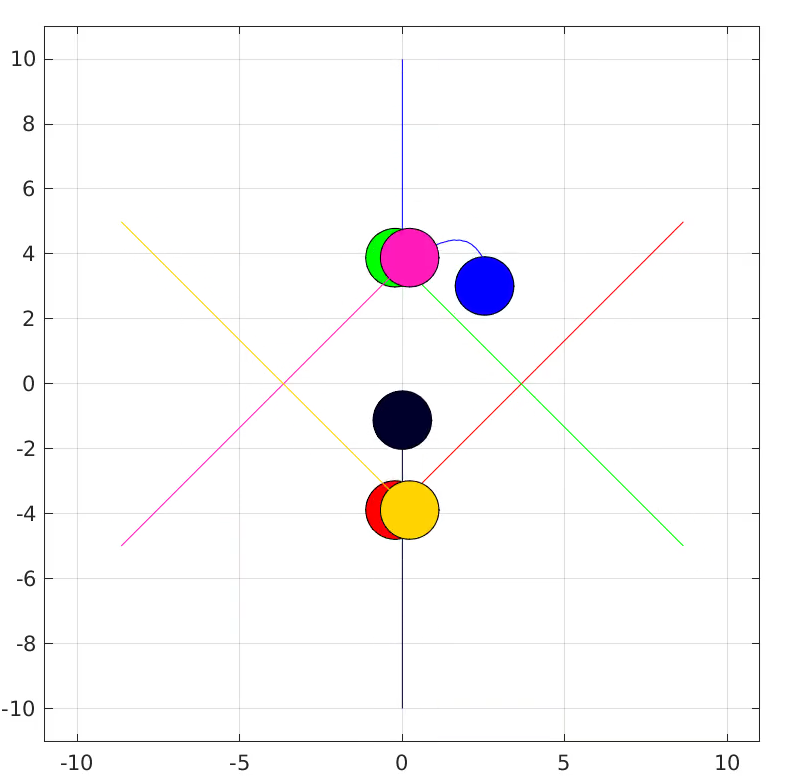}
     \end{subfigure}\label{SA:c}
     \begin{subfigure}[b]{0.5\linewidth}
         \centering
         \includegraphics[width=\textwidth]{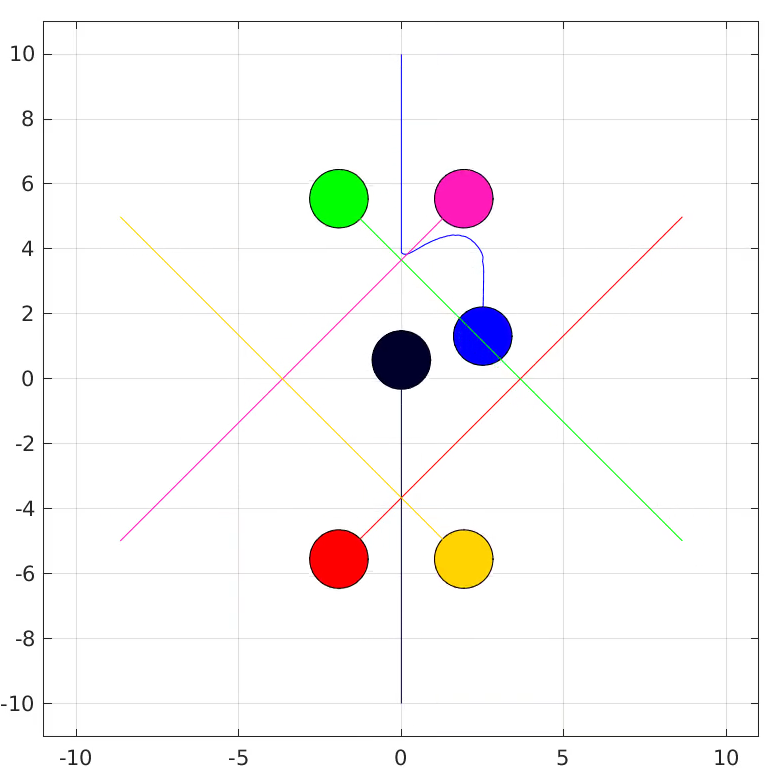}
     \end{subfigure}\label{SA:d}
     \vspace{0.1cm}
     \hspace{-0.2cm}
     \begin{subfigure}[b]{0.5\linewidth}
         \centering
         \includegraphics[width=\textwidth]{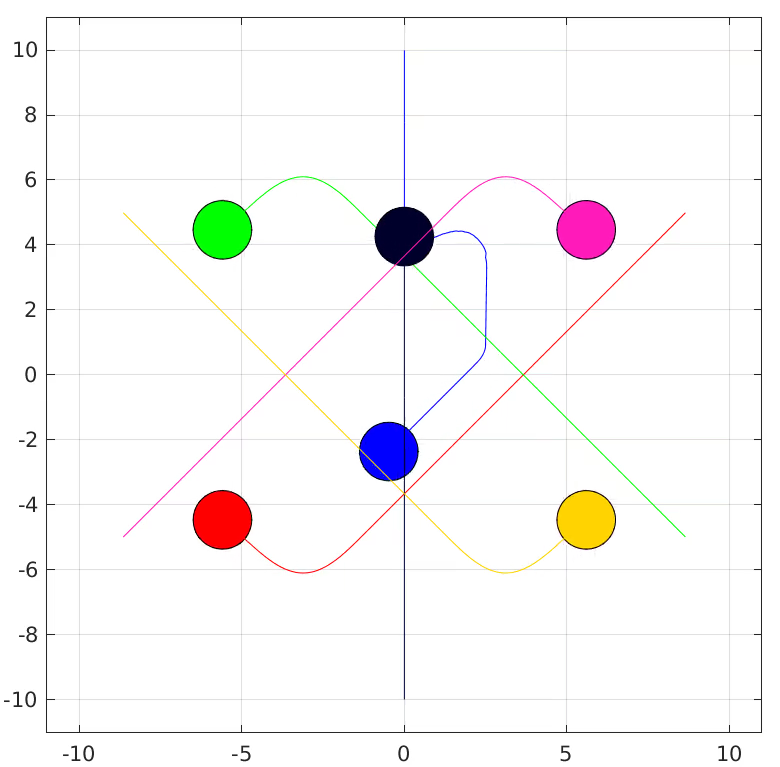}
     \end{subfigure}\label{SA:e}
     \begin{subfigure}[b]{0.5\linewidth}
         \centering
         \includegraphics[width=\textwidth]{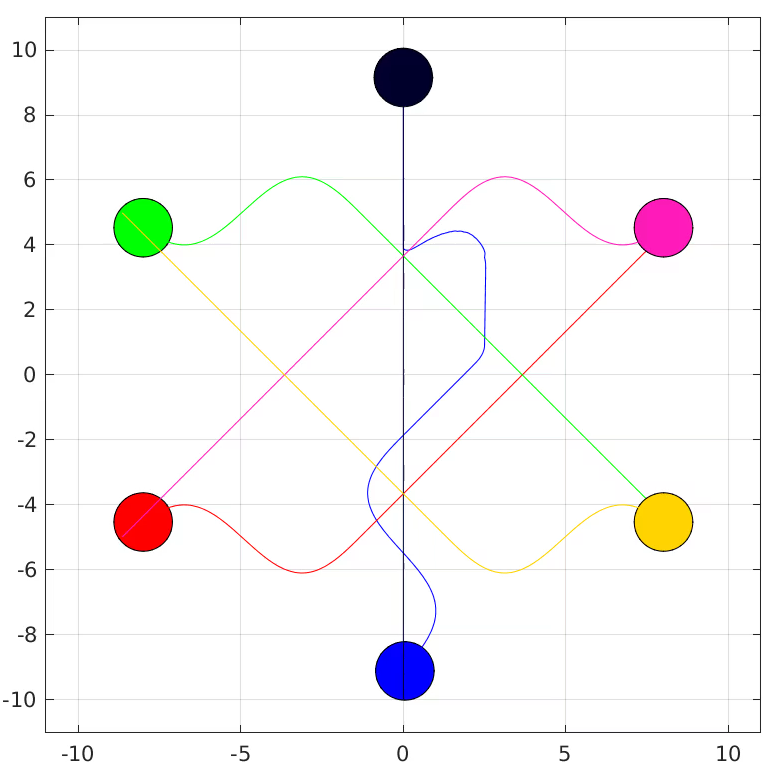}
     \end{subfigure}\label{SA:f}
        \caption{Blue disc represents the agent while rest are the obstacles with simple behaviour}
        \label{fig:single-agent}
\end{figure}

First we validate our formulation in a single agent case. Figure (\ref{fig:single-agent}) shows the scenario where single agent is among five dynamic obstacles. All the participants in the environment are of same radius and have same speed limits. As can be seen the agent executes safe trajectories to avoid all the obstacles and reaches the goal. The computation time for each cycle in this scenario is around 10ms making it achieve an update rate of 100Hz.

\subsection{Multiple agents}\label{section:results-multiagent}

\begin{figure}
     \centering
     \hspace{-0.2cm}
     \begin{subfigure}[b]{0.5\linewidth}
         \centering
         \includegraphics[width=\textwidth]{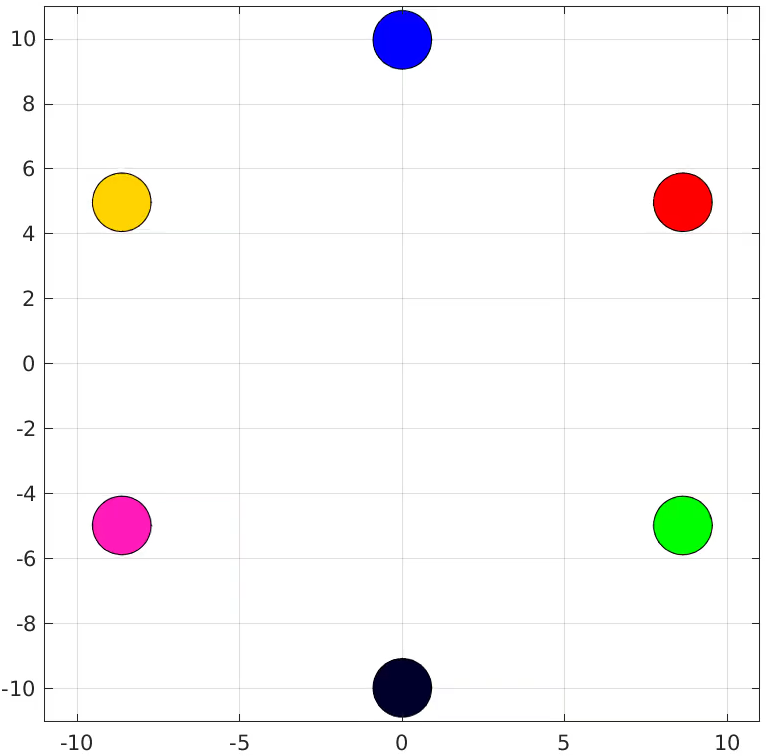}
     \end{subfigure}
     \begin{subfigure}[b]{0.5\linewidth}
         \centering
         \includegraphics[width=\textwidth]{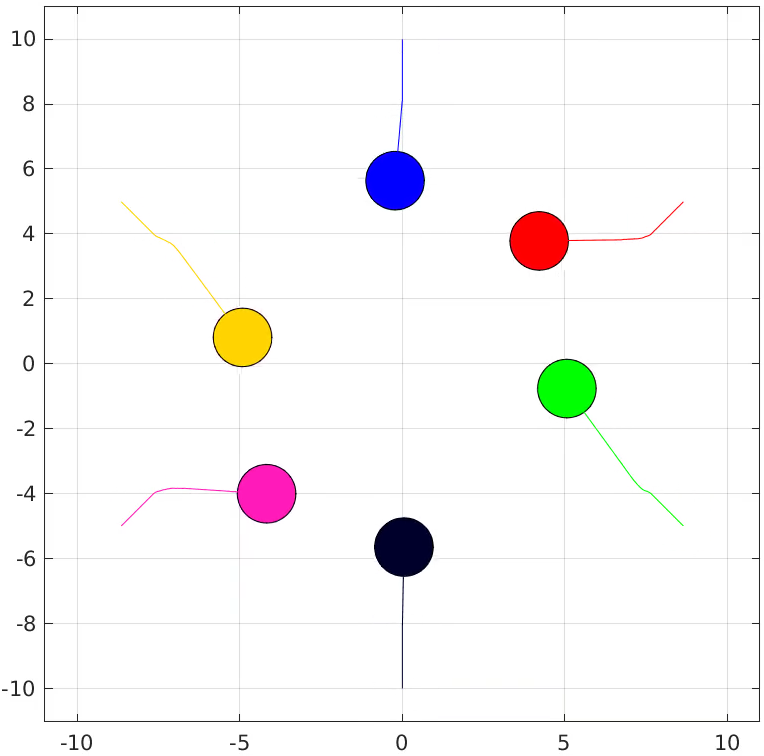}
     \end{subfigure}
     \vspace{0.1cm}
     \hspace{-0.2cm}
     \begin{subfigure}[b]{0.5\linewidth}
         \centering
         \includegraphics[width=\textwidth]{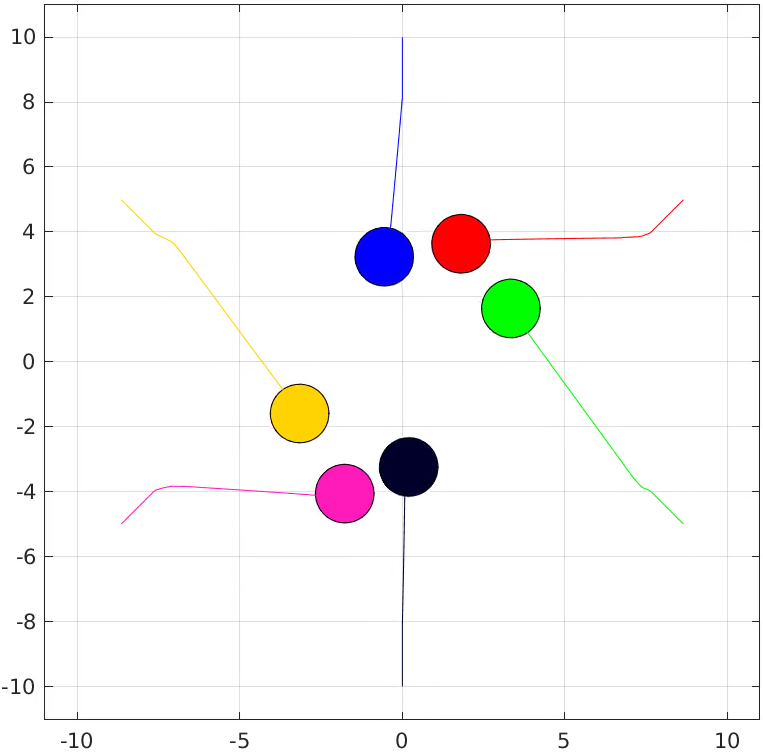}
     \end{subfigure}
     \begin{subfigure}[b]{0.5\linewidth}
         \centering
         \includegraphics[width=\textwidth]{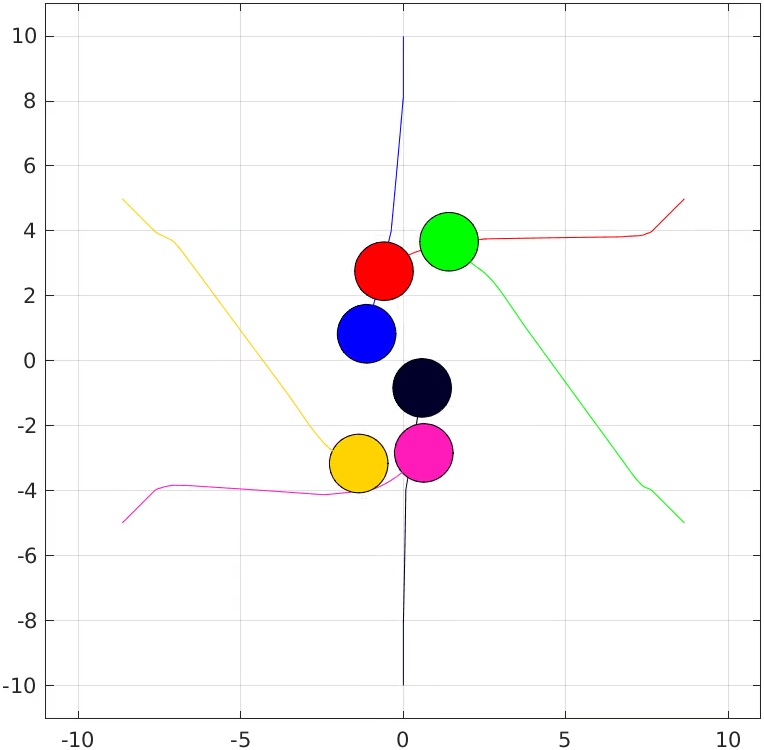}
     \end{subfigure}
     \vspace{0.1cm}
     \hspace{-0.2cm}
     \begin{subfigure}[b]{0.5\linewidth}
         \centering
         \includegraphics[width=\textwidth]{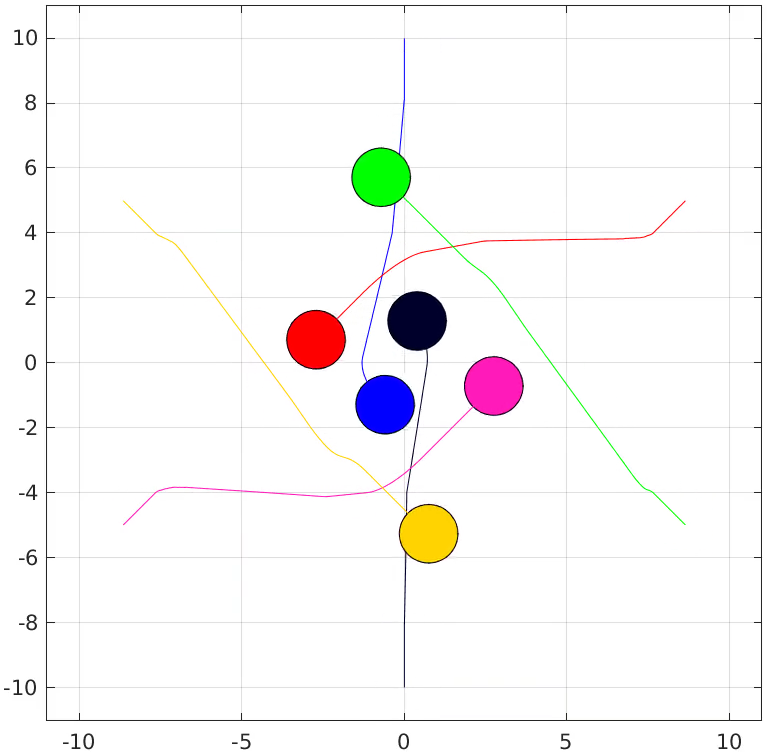}
     \end{subfigure}
     \begin{subfigure}[b]{0.5\linewidth}
         \centering
         \includegraphics[width=\textwidth]{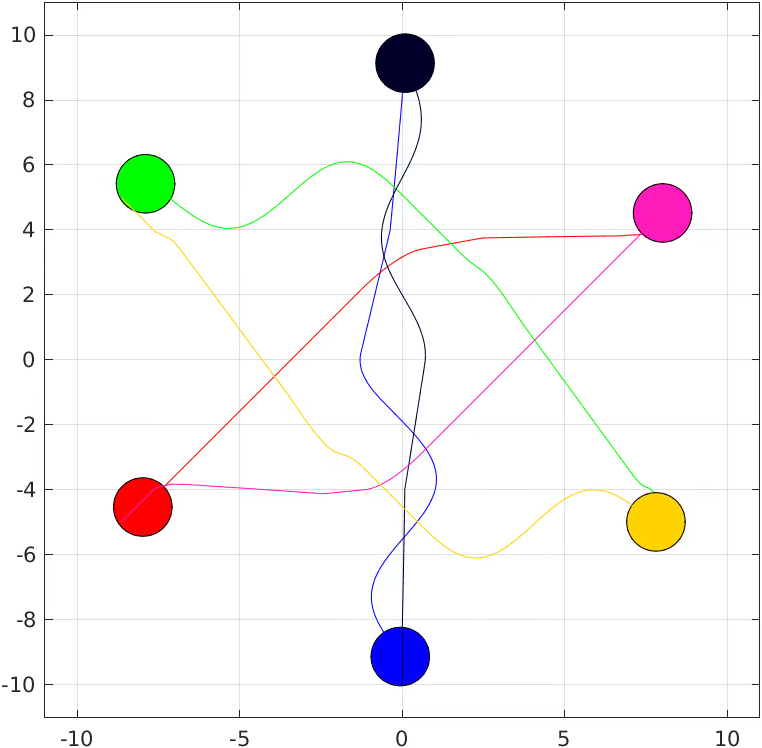}
     \end{subfigure}
        \caption{Multi agent scenario: 6 agents}
        \label{fig:6-agents}
\end{figure}

In this section, we evaluated the performance of our Inverse Velocity Obstacles in a multi-agent collision scenario. We first evaluate for a 6 agent scenario in an antipodal case. All the agents are of same radius and have same speed and acceleration limits. Figure \ref{fig:6-agents} shows the scenario. All the agents plan independently considering all the other participants as potential obstacles. As can be seen all the agents generate safe motions avoiding each other and reach the goal. The computational time for each cycle in this scenario is 15ms with update rates of 66Hz.

\begin{figure}
     \centering
     \hspace{-0.2cm}
     \begin{subfigure}[b]{0.5\linewidth}
         \centering
         \includegraphics[width=\textwidth]{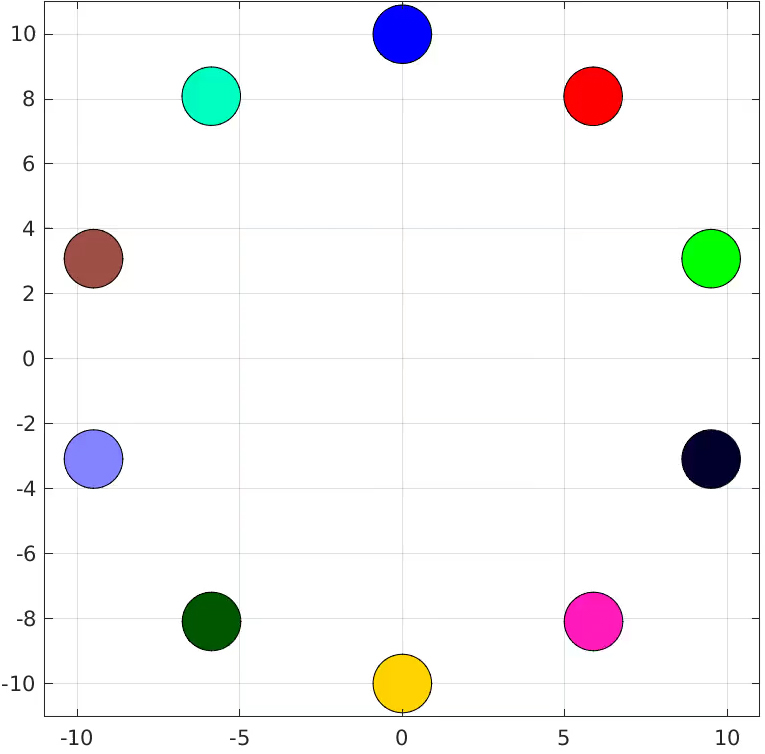}
         \caption{}
         \label{MA:a}
     \end{subfigure}
     \begin{subfigure}[b]{0.505\linewidth}
         \centering
         \includegraphics[width=\textwidth]{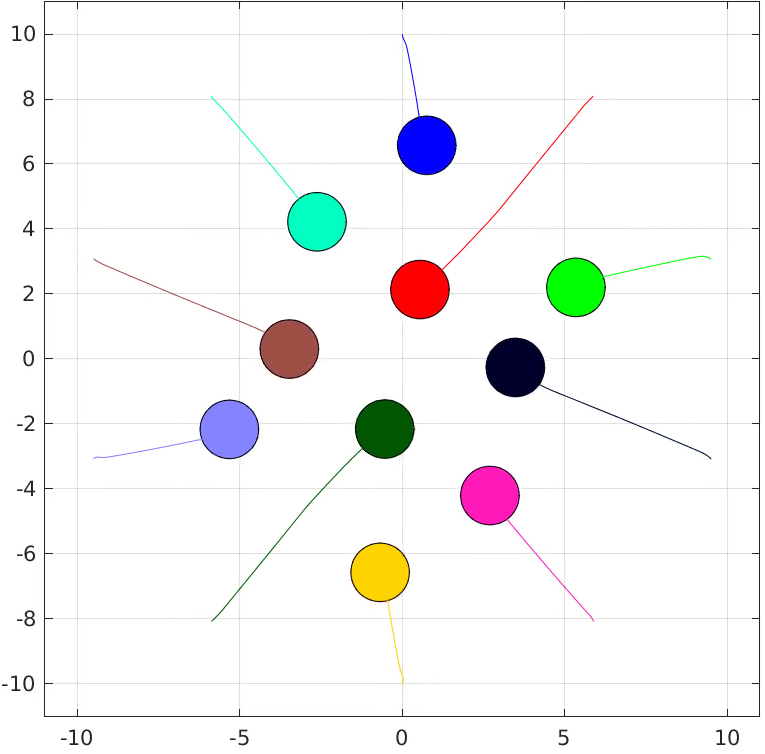}
         \caption{}
         \label{MA:b}
     \end{subfigure}
     \vspace{0.1cm}
     \hspace{-0.2cm}
     \begin{subfigure}[b]{0.5\linewidth}
         \centering
         \includegraphics[width=\textwidth]{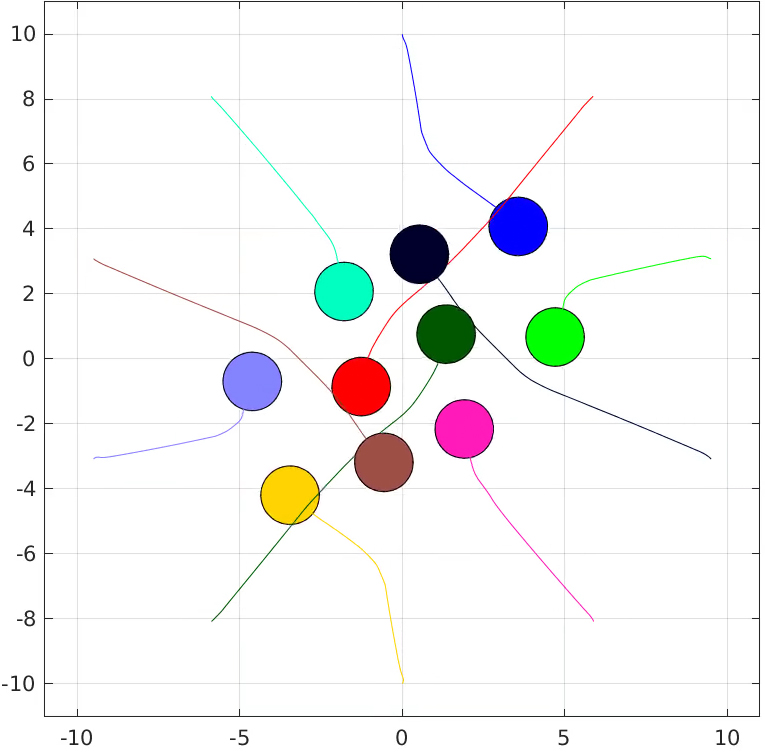}
         \caption{}
         \label{MA:c}
     \end{subfigure}
     \begin{subfigure}[b]{0.5\linewidth}
         \centering
         \includegraphics[width=\textwidth]{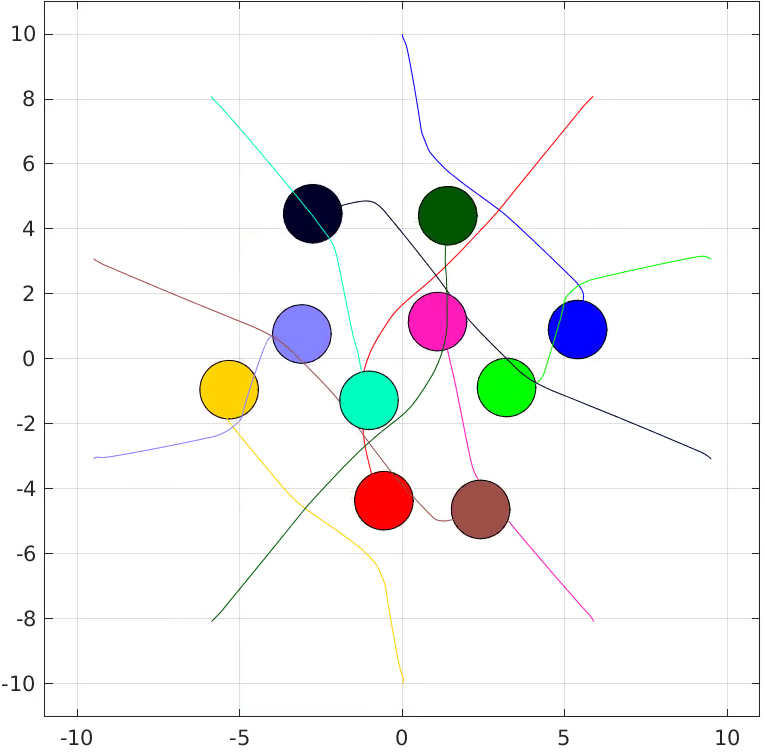}
         \caption{}
         \label{MA:d}
     \end{subfigure}
     \vspace{0.1cm}
     \hspace{-0.2cm}
     \begin{subfigure}[b]{0.5\linewidth}
         \centering
         \includegraphics[width=\textwidth]{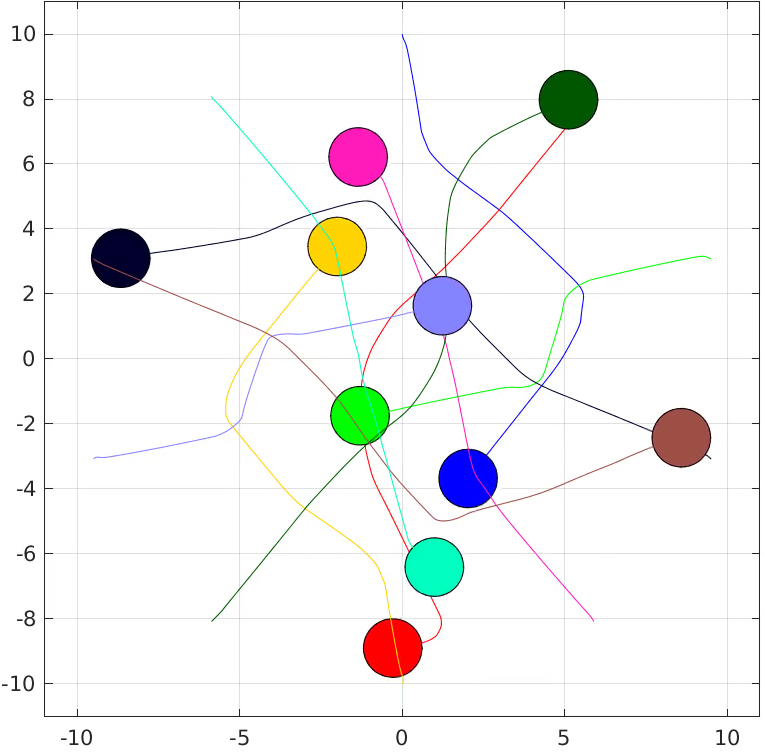}
         \caption{}
         \label{MA:e}
     \end{subfigure}
     \begin{subfigure}[b]{0.5\linewidth}
         \centering
         \includegraphics[width=\textwidth]{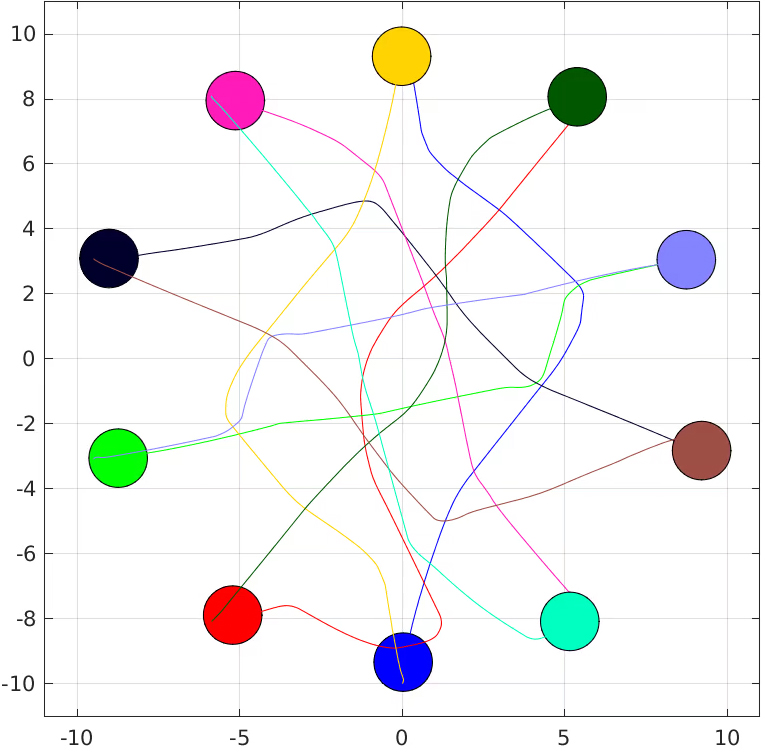}
         \caption{}
         \label{MA:f}
     \end{subfigure}
        \caption{Multiagent scenario: 10 agents}
        \label{fig:10-agents}
\end{figure}

Next, we increased the number of agents in the same scenario with same settings to validate how IVO scales when the agents grow. Figure \ref{fig:10-agents} presents the scenario with 10 agents that is evaluated. The computational time increases with the increase in the number of agents and for this scenario it is around 15ms for each cycle and has the update rates close to 50Hz. Even though the computational time is increasing with the increase in the number of agents, the update rates are high enough for aiding in a easy real time implementation.

Additional simulation results are available at \href{https://sites.google.com/view/inverse-velocity-obstacle}{https://sites.google-.com/view/inverse-velocity-obstacle}.

\subsection{Real time Experiments}

\begin{figure}
     \centering
     \hspace{-0.2cm}
     \begin{subfigure}[b]{0.5\linewidth}
         \centering
         \includegraphics[width=\textwidth]{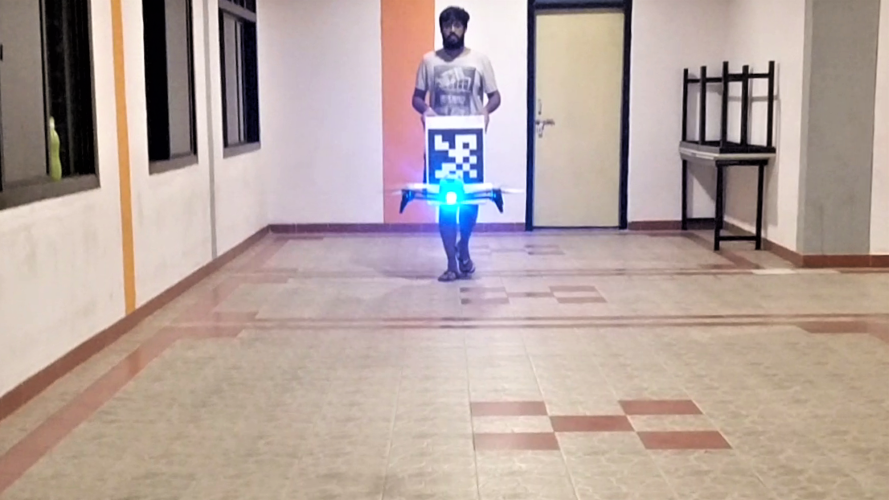}
         \caption{}
         \label{real:a}
     \end{subfigure}
     \begin{subfigure}[b]{0.505\linewidth}
         \centering
         \includegraphics[width=\textwidth]{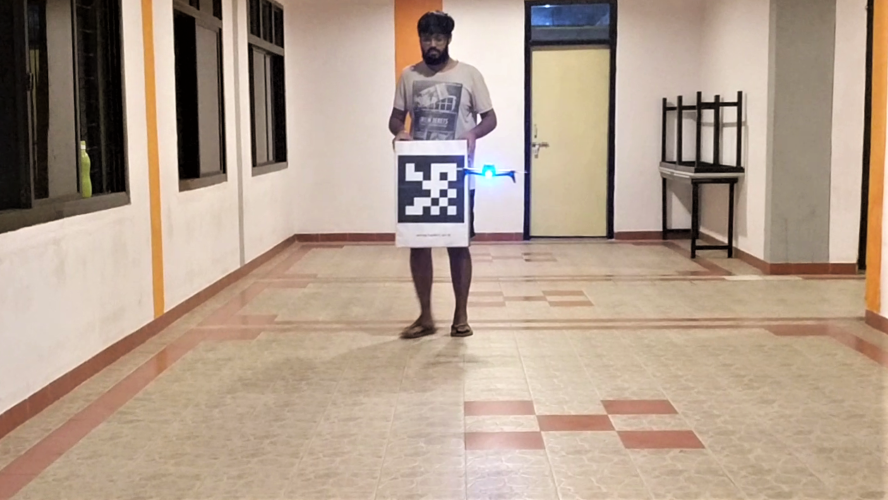}
         \caption{}
         \label{real:b}
     \end{subfigure}
     \vspace{0.1cm}
     \hspace{-0.2cm}
     \begin{subfigure}[b]{0.5\linewidth}
         \centering
         \includegraphics[width=\textwidth]{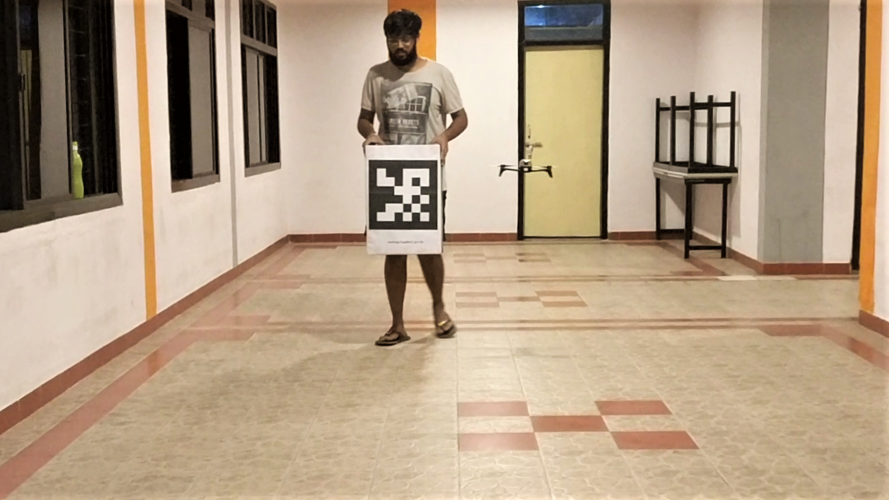}
         \caption{}
         \label{real:c}
     \end{subfigure}
     \begin{subfigure}[b]{0.5\linewidth}
         \centering
         \includegraphics[width=\textwidth]{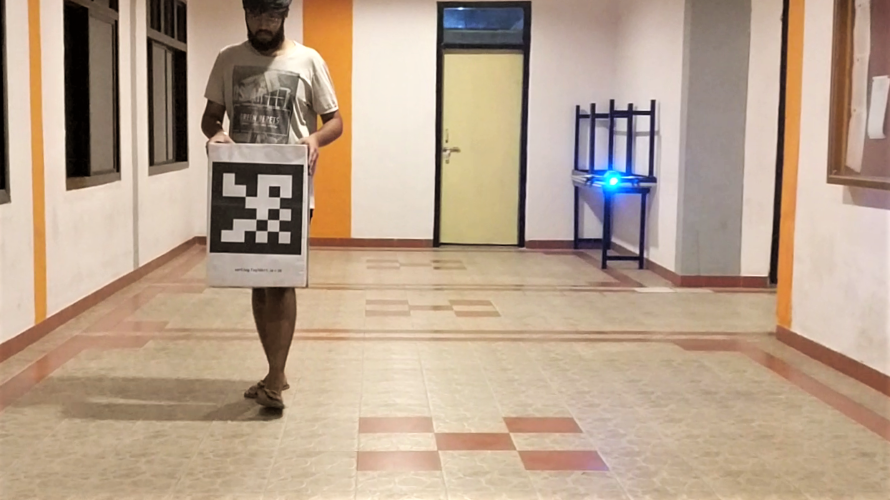}
         \caption{}
         \label{real:d}
     \end{subfigure}
     \vspace{0.1cm}
     \hspace{-0.2cm}
     \begin{subfigure}[b]{0.5\linewidth}
         \centering
         \includegraphics[width=\textwidth]{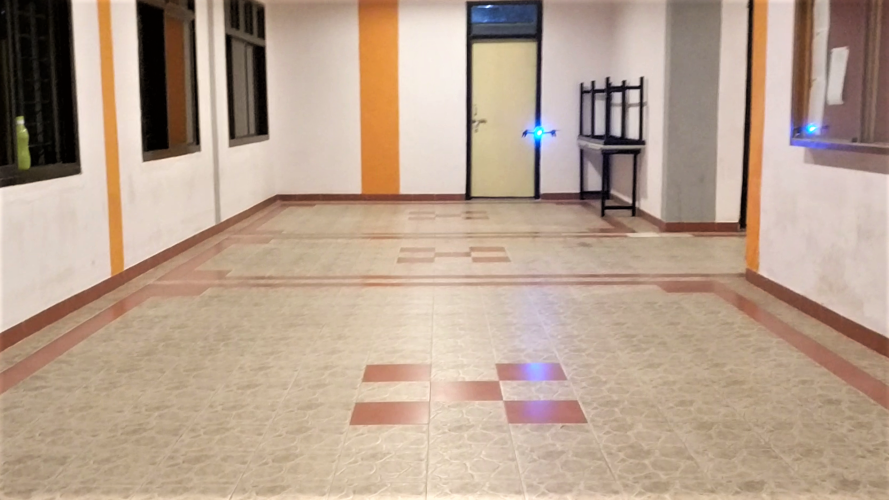}
         \caption{}
         \label{real:e}
     \end{subfigure}
     \begin{subfigure}[b]{0.5\linewidth}
         \centering
         \includegraphics[width=\textwidth]{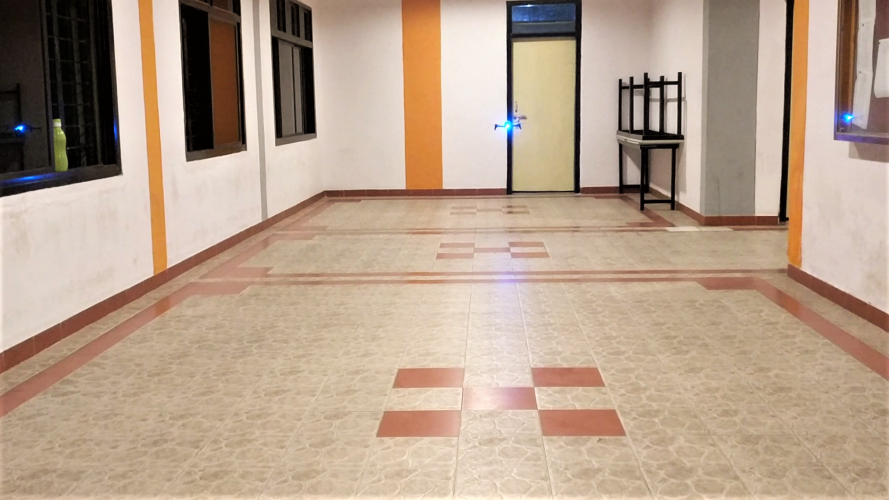}
         \caption{}
         \label{real:f}
     \end{subfigure}
        \caption{Real time implementation with one dynamic obstacle}
        \label{fig:single-agent-real}
\end{figure}

In this section, we evaluate the performance of Inverse Velocity Obstacles in real time implementation. For this we used Parrot Bebop2 quadrotor which accepts the yaw, pitch, roll angles as the control input. We have also developed a PID controller for the velocity control. This is integrated on top of the inbuilt controller for better performance for the validation of the algorithm as our algorithm is developed in velocity control space.  This lets us pass velocity as a control command to the drone. We used April Tags \cite{olson2011tags} of the family Tag36h11 for better state estimation of the other participants in the environment. We have completely bypassed the self state estimation module as our framework does not need the agents self state for collision detection and avoidance. Figures(\ref{real:a})-(\ref{real:f}) show the snapshots of the real time implementation of the proposed method on the quadrotor in a dynamic environment.

\subsection{Comparisons with Velocity Obstacle for Collision Detection}

In this section we compare the presented approach with Velocity Obstacle and show that the collision detection for IVO is more reliable compared to the traditional Velocity Obstacle. For this the equation \ref{eq:vo}, is re-written in terms of controls in the following manner,
\begin{equation}
    \label{eq:vel-cone-alg}
    f = c_1\dot{x_r}^2 + c_2\dot{y_r}^2 + c_3\dot{x_r}\dot{y_r} + c_4\dot{x_r} + c_5\dot{y_r} + c_6
\end{equation}

Similarly, the original Velocity Obstacle equation is rearranged  into equation \ref{eq:vel-cone-alg}. In a real time scenario, the coefficients $c_i$ take the form of a random variable. This introduces randomness into each coefficient due to the uncertainties in the state, actuation and perception.

\begin{equation*}
    c_i = \alpha P_i(x_r, y_r, x_o, y_o, \dot{x_o}, \dot{y_o})
\end{equation*}

\begin{figure}
     \centering
     \hspace{-0.15cm}
     \begin{subfigure}[b]{0.5\linewidth}
         \centering
         \includegraphics[width=\textwidth]{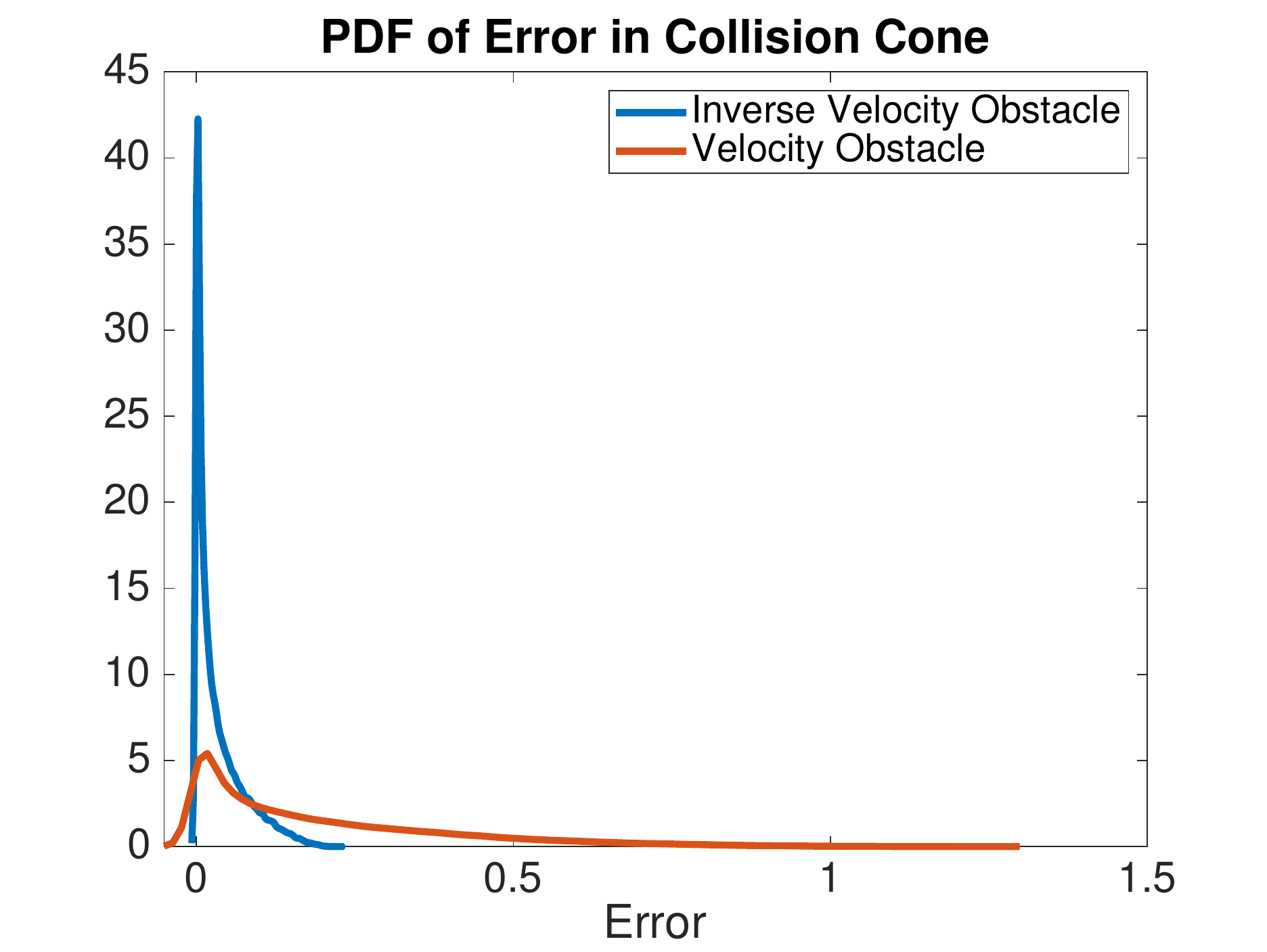}
     \end{subfigure}
     \begin{subfigure}[b]{0.5\linewidth}
         \centering
         \includegraphics[width=\textwidth]{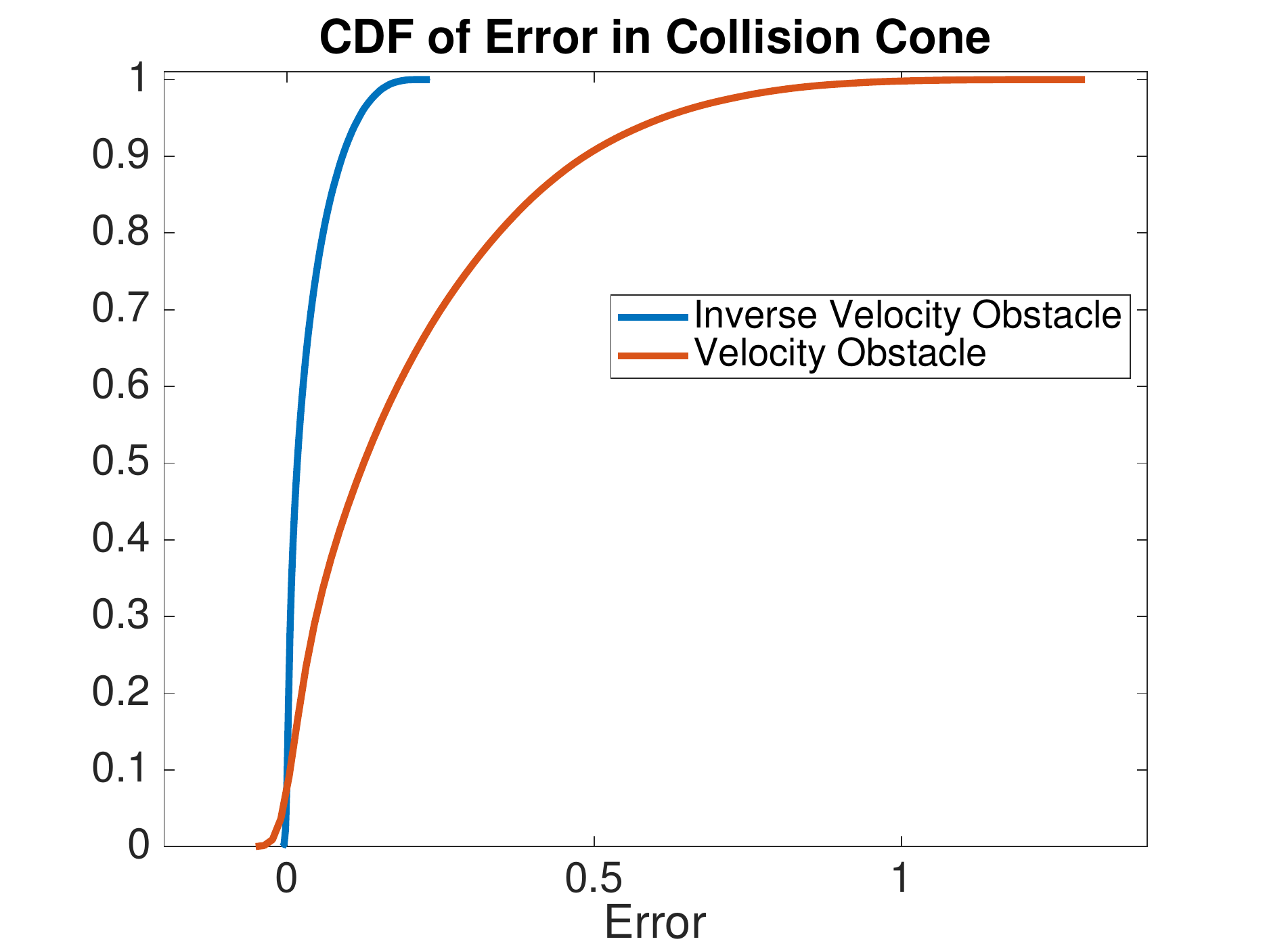}
     \end{subfigure}
        \caption{Distributions for collision cone}
        \label{fig:pdf-cdf}
\end{figure}

$P_i(.)$ denotes the PDF of $c_i$. The advantage with IVO is that the random variables need not depend on $x_r$ and $y_r$. In figure \ref{fig:pdf-cdf}, we compare the probability distributions of the error in collision cone for velocity obstacle as well as inverse velocity obstacle. The noise in agent and obstacle states were assumed to be Gaussian distributions with zero mean. The distributions clearly show a reduction in the noise. The 99\% confidence region for inverse velocity obstacle is between 0 and 0.14 error range while it is between -0.03 and 0.56 error range for velocity obstacle. This provides a better scope for dealing with the noise just by increasing the radius of the obstacle.

\begin{figure}
    \centering
    \includegraphics[page=1,width=\linewidth]{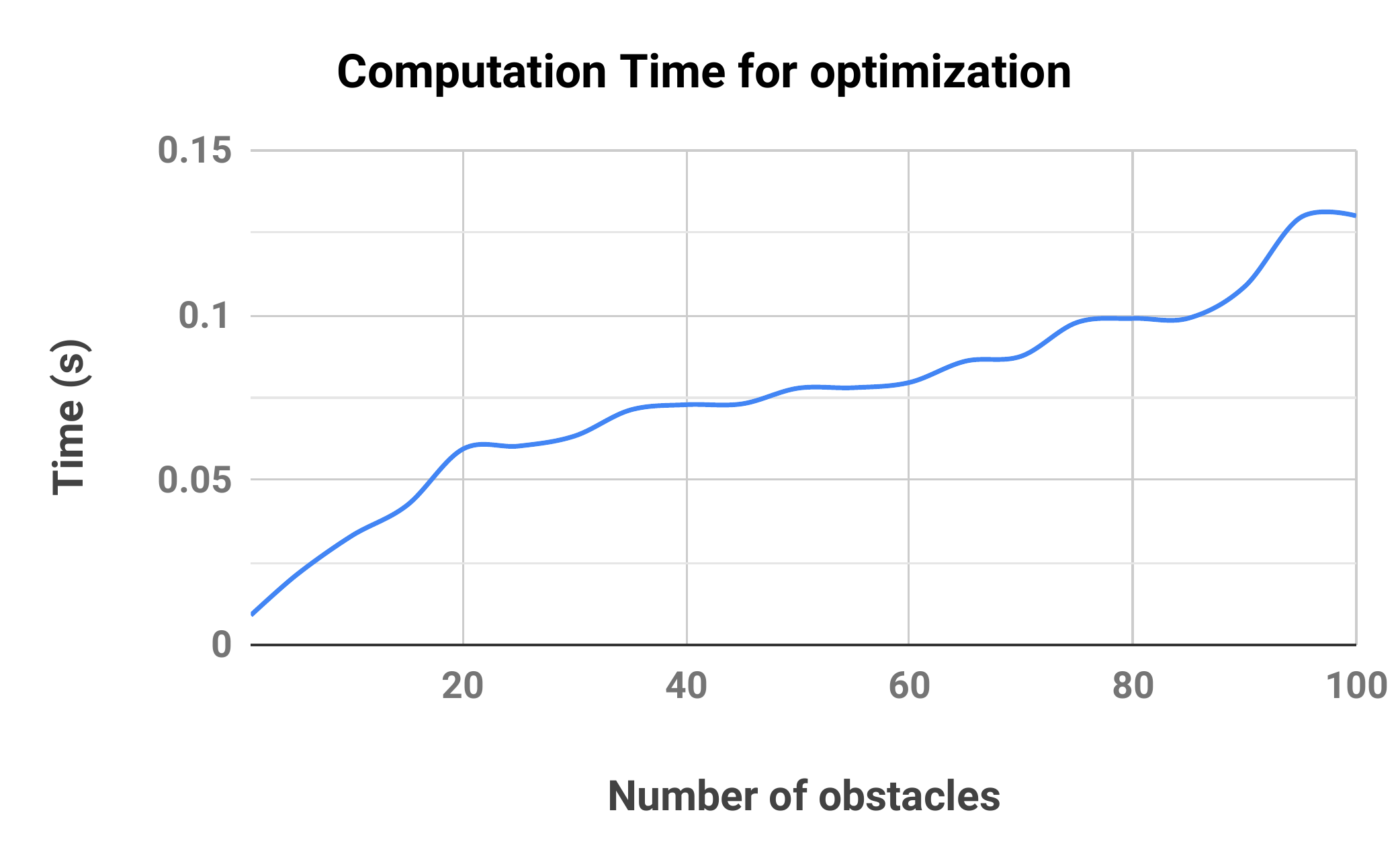}
    \caption{Computational time for different number of obstacles}
    \label{fig:computational-time}
\end{figure}

\section{Conclusion}\label{concl}
In this paper, we presented a new concept called Inverse Velocity Obstacles, for the safe navigation of autonomous agents in dynamic environments. In contrast to the previous works, we developed an ego-centric framework which eliminates the reliance on robot's state for collision detection. This also decreases the computational complexity improving the real time implementation. The formulation presented is a natural extension of Velocity Obstacle and is easy to implement. We have also applied this to multi-agent navigation and we show its efficacy to generate natural paths for systems as high as 50 agents in very tight environment.  

Our further work includes investigating the Inverse Velocity Obstacle application in the domains like crowd simulations and rescue works. Also we are exploring to extending the method to handle non-parametric uncertainty that arises due to perception and localization error.

\bibliographystyle{unsrt}
\bibliography{air} 

\end{document}